\documentclass[11pt]{article}

\usepackage{amsmath}
\usepackage{enumitem}
\usepackage{booktabs}

\usepackage[final]{acl}

\usepackage{times}
\usepackage{latexsym}

\usepackage[T1]{fontenc}

\usepackage[utf8]{inputenc}

\usepackage{microtype}

\usepackage{inconsolata}

\usepackage{graphicx}
\usepackage{hyperref}       
\usepackage{url}            
\usepackage{amsfonts}       
\usepackage{nicefrac}       
\usepackage{xcolor}         
\usepackage{adjustbox}
\usepackage{multirow}
\usepackage[most]{tcolorbox}

\newcommand{\tactics}{\textsc{TACTICS}}
\newcommand{\dsir}{\textsc{DSIR}}
\newcommand{\clusterclip}{ClusterCLIP}

\newcommand{\awstranslate}{NMT\textsubscript{base}}
\newcommand{\llmmt}{LLM\textsubscript{MT}}

\newcommand{\covtwenty}{Cov$_{20}$}

\newcommand{\mt}{MT}
\newcommand{\mqm}{MQM}

\title{TACTICS: Taxonomy-Aware Intelligent Corpus Sampling for Machine Translation}

\author{
  \textbf{Prasanth Bathala},
  \textbf{Anubhav Shrimal},
  \textbf{Sukhdeep Singh Kharbanda},
\\
  \textbf{Pradyumna Lanka},
  \textbf{Rohit Dhaipule}
\\
  Translation Services, Amazon
\\
  \small{
    \{pbathala, shrimaa, kharsukh, pradyl, rdhaipu\}@amazon.com
  }
}

\begin{document}
\maketitle
\begin{abstract}
Large-scale machine-translation (MT) systems are typically evaluated on
random samples from a corpus whose distributional composition is an artifact
of how it was assembled. Such a sample inherits the phenomena the collection
happens to contain rather than the full space a system must handle, spanning
rule-governed conventions (terminology, punctuation, currency formatting) and
context-dependent phenomena (tone, honorifics, document-level coherence), and
thus provides no coverage guarantee for assessing robustness. We propose
\textbf{TACTICS} (\textbf{T}axonomy-\textbf{A}ware
\textbf{C}overage-op\textbf{T}imized \textbf{I}ntelligent \textbf{C}orpus
\textbf{S}ampling), which recasts coverage as an explicit objective. TACTICS
induces a hierarchical taxonomy from a locale style guide,
classifies segments against it, and selects a fixed-budget subset jointly
optimizing coverage of rare categories, document-level coherence, and
distributional fidelity to the full corpus. Applied to MT evaluation across
four translation directions, TACTICS improves coverage of rare categories
over lexical and embedding-based selection. By targeting the phenomena that
separate systems, TACTICS makes a fixed evaluation budget go further,
recovering the true system ranking from far fewer segments than random
sampling wherever a real quality gap exists and never signaling a difference
where none exists.
\end{abstract}

\section{Introduction}
\label{sec:introduction}

Modern machine translation (\mt{}) systems, including both dedicated neural
MT engines and LLM-based translation pipelines, are increasingly deployed at
scale across many language pairs and locales
\citep{bahdanau2015neural,vaswani2017attention,brown2020language,touvron2023llama,openai2023gpt4}.
At this scale, evaluation is a bottleneck. Human assessment of every
translated segment is prohibitively expensive, so systems are typically
evaluated on a random sample drawn from a large corpus. The implicit
assumption is that such a sample provides a representative picture of
system quality. But a representative sample is not the same as a diagnostic
one. A random sample reproduces the corpus faithfully, yet provides no
guarantee of adequate coverage of the phenomena that discriminate between
systems.

The difficulty is that these discriminating phenomena are also the ones a
corpus represents least evenly. Most of a translation corpus consists of short, common segments that nearly any
system translates correctly, whereas the cases that genuinely stress a system,
such as terminology constraints, locale-specific formatting, and date or currency
conventions, are comparatively rare. This
matters because an evaluation, whether by automatic metric~\citep{sai2022survey}
or human protocol such as \mqm{}~\citep{lommel2014multidimensional}, is only as
diagnostic as the set on which it is computed. When easy segments dominate
that set, the differences concentrated in the harder ones are averaged away,
and two systems of unequal quality can appear all but indistinguishable.

Existing selection methods do not close this coverage gap. Most target
training-data curation rather than evaluation, whether through quality
filtering \citep{gao2021pile,marion2023less,tirumala2023d4}, importance
resampling on lexical or n-gram statistics such as \dsir{}
\citep{xie2023dataselectionlanguagemodels}, or embedding-space diversity such
as \clusterclip{} \citep{shao2024balanceddatasamplinglanguage}. These signals
capture frequent lexical patterns or broad semantic variation, but none of
them explicitly models the prescriptive rules that govern localization,
such as whether a brand term is preserved, punctuation follows locale
conventions, or currency and dates are formatted correctly. A sample can
therefore be lexically or semantically diverse and still fail to cover the
phenomena on which systems actually differ.

\begin{figure*}[t]
    \centering
    \includegraphics[width=\textwidth, trim=45 615 50 55, clip]{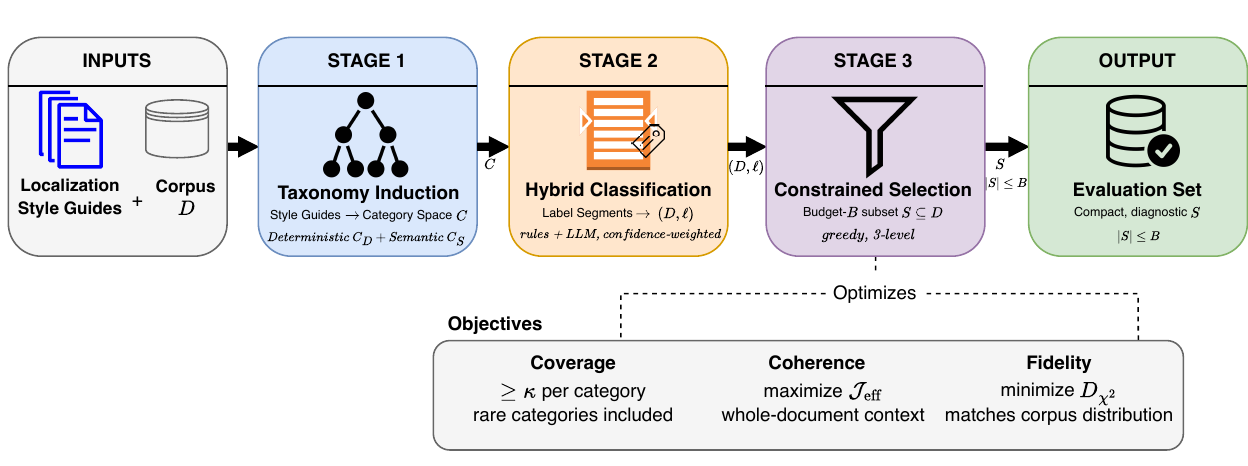}
    \caption{
    Overview of \tactics{}. Stage~1 induces a taxonomy (category space $C$)
    from style guides; Stage~2 labels segments into $C$, yielding $(D,\ell)$;
    Stage~3 selects a budget-$B$ subset $S$ jointly optimizing coverage
    ($\ge\kappa$), coherence ($\mathcal{J}_{\mathrm{eff}}$), and fidelity
    ($D_{\chi^2}$).
    }
    \label{fig:tactics_framework}
\end{figure*}

We propose \textbf{\tactics{}} (\textbf{T}axonomy-\textbf{A}ware
\textbf{C}overage-op\textbf{T}imized \textbf{I}ntelligent
\textbf{C}orpus \textbf{S}ampling), a data-selection method that constructs
compact, stratified evaluation sets by treating coverage as an explicit
objective rather than a byproduct of how the corpus was sampled. It
formulates evaluation-set construction as a constrained optimization problem
over a hierarchical taxonomy induced from prescriptive style guides. We
instantiate this for \mt{}, though the approach applies to any evaluation
that must cover a structured set of phenomena under a fixed budget.
\tactics{} balances three properties (Figure~\ref{fig:tactics_framework}).
\begin{enumerate}[label=(\roman*), leftmargin=*, itemsep=1pt, topsep=2pt, parsep=0pt]
    \item \textbf{Coverage.} Every taxonomy category, including rare ones, is
    adequately represented under a fixed budget, unlike random sampling.
    \item \textbf{Coherence.} Segments are selected with their document
    neighbors, so each is evaluated in context rather than in isolation.
    \item \textbf{Fidelity.} The sample matches the full corpus in category
    composition and difficulty.
\end{enumerate}

We evaluate \tactics{} on a large-scale localization corpus across four
translation directions. At a fixed budget of $B=5{,}000$,
\tactics{} raises the average number of taxonomy categories with at least 20
sampled segments from 56.3 to 73.3 relative to random sampling, while
preserving high correlation with the full-corpus category distribution
(Pearson $r=0.949$ to $0.982$) and close alignment on edit rate. It also
increases the average segments per document from 1.9 to 3.2, yielding less
fragmented samples. Most importantly, these subsets are more diagnostic. Where a real quality gap
exists, \tactics{} recovers the true system ranking from fewer segments than
random, with a higher paired $t$-statistic at every budget on
en$\rightarrow$hi, and raises no false signal on directions where the systems
are tied.

Our contributions are:
\begin{itemize}[leftmargin=3.0em, itemsep=2pt, topsep=2pt, parsep=0pt]
    \item We introduce a \textbf{style-guide-grounded taxonomy induction pipeline}
    for representing localization phenomena relevant to \mt{} evaluation.
    \item We formulate fixed-budget evaluation set construction as a
    \textbf{taxonomy-aware constrained sampling problem} over coverage,
    coherence, and fidelity.
    \item We show that \tactics{} \textbf{covers rare phenomena} and
    \textbf{recovers true system quality differences} more reliably than
    random, lexical, and embedding-based selection.
\end{itemize}

\section{Related Work}
\label{sec:related}

\subsection{MT Evaluation and Test Set Design}

MT evaluation relies on automatic metrics such as
BLEU~\citep{bleu2002}, TER~\citep{snover2006study},
chrF~\citep{popovic-2015-chrf}, COMET~\citep{rei2020comet}, and
BLEURT~\citep{sellam2020bleurtlearningrobustmetrics}, whose reliability for
ranking systems has been studied at scale~\citep{kocmi2021ship}, alongside
human frameworks such as MQM~\citep{lommel2014multidimensional,freitag2022mqm}.
These estimate quality on a test set but not whether the set is diagnostic,
that is, whether it contains the phenomena on which systems differ. This depends on how the
set is built. The WMT shared task builds its test sets by drawing source texts from a chosen
domain, usually online news, and commissioning reference translations for the
task, leaving targeted phenomenon evaluation to separately contributed test
suites~\citep{barrault2020findings}. Its composition is therefore fixed by the
choice of domain and source rather than by explicit coverage of these phenomena,
and reproducing it for a new language pair or domain requires substantial manual
effort. Benchmarks like FLORES-101 and FLORES-200 scale instead by sampling
their sentences from a corpus such as
Wikipedia~\citep{goyal2022flores101,nllb2022}, but inherit whatever the
corpus contains, with no guarantee that the discriminating phenomena are covered. \tactics{} addresses both limitations by making
coverage of a style-guide-derived taxonomy an explicit objective when sampling a
corpus, so those phenomena that discriminate systems are represented by
construction rather than by chance or manual curation.

\subsection{Data Selection and Sampling}

Prior data-selection work has largely focused on training data, including
quality filtering~\citep{gao2021pile,marion2023less}, pruning~\citep{tirumala2023d4}, importance resampling such as \dsir{}~\citep{xie2023dataselectionlanguagemodels}, and embedding-based
diversity such as \clusterclip{}~\citep{shao2024balanceddatasamplinglanguage}. These methods
select examples using lexical statistics, embeddings, or training utility.

A separate line of work selects data for \emph{evaluation} rather than training,
so that systems can be tested on fewer examples. One group reduces a benchmark
to a small subset that still reproduces its system ranking, keeping the examples
that models find hardest or most
discriminating~\citep{vivek2024anchor,polo2024tinybenchmarks,rodriguez2021evaluation,vania2021comparing,perlitz2024efficient}.
Another selects examples adaptively, spending human judgments on the comparisons
that best separate systems~\citep{chaganty2018price,mohankumar2022active,kossen2021active}.
More recently, \citet{zouhar2025select} pick segments for human evaluation of NLG
systems from metric-score variance, output diversity, and item difficulty. All
of these rely on signals read from model outputs. The selection is therefore
tied to the systems being evaluated, does not transfer easily to new ones, and
cannot ensure that the phenomena a system should handle are covered.

\tactics{} inverts this dependence. We cast evaluation-set construction as
taxonomy-aware stratified sampling under a fixed budget, with strata drawn from a
prescriptive style-guide taxonomy rather than surface metadata, domains, or
embedding clusters. The taxonomy is fixed independently of any model, so the resulting evaluation
set is reusable across systems, including ones built later. This approach suits
MT, where much of localization quality depends not on meaning but on explicit
rules, such as preserving brand terms, formatting dates and currencies for the
locale, or choosing the correct honorific. These rules are set by style guides,
yet selection by lexical or embedding similarity fails to capture them
(Section~\ref{sec:results}).

\subsection{Automated Taxonomy Construction}

Recent LLM-based taxonomy induction methods build taxonomies bottom-up from
observed data, e.g., TnT-LLM~\citep{wan2024tntllmtextminingscale} and
Chain-of-Layer~\citep{zeng2024chainoflayeriterativelypromptinglarge}. Such
methods discover corpus-salient patterns, but localization evaluation also
depends on prescriptive rules that may be scarce in a given corpus, such as
currency formatting, brand preservation, quotation marks, or date
conventions. \tactics{} takes a complementary top-down approach. It induces
taxonomy structure from localization style guides and grounds it in the
corpus through constrained classification and sampling, so evaluation sets
cover both common patterns and rare style-guide-defined phenomena that matter
for system comparison.

\section{Methodology}

\subsection{Problem Formulation}
\label{sec:problem_formulation}

Let $D=\{s_i\}_{i=1}^n$ denote translation segments, with each
segment belonging to a document $J \in \mathcal{J}$. Let $T$ be a
hierarchical taxonomy with category set $C=C_D\cup C_S$, where
deterministic categories $C_D$ are rule-matchable and semantic categories
$C_S$ are LLM-inferred. Each segment has taxonomy labels
$\ell(s)\subseteq C$ and a post-edit score $e(s)\in[0,1]$, the fraction of a
machine translation that a human corrected (approximated by TER). The score can
come from any MT system and serves only to approximate the segment's difficulty.

Given a budget $B\ll |D|$, we select a stratified evaluation subset
$S\subset D$, $|S|\leq B$, by enforcing coverage constraints and optimizing
the coherence and fidelity objectives below.

\textbf{Coverage (constraint).}
For a minimum threshold $\kappa \ge 1$, each category with at least $\kappa$
segments in the corpus must be represented in the sample:
\begin{equation}\label{eq:coverage}
  \forall c \in C:\quad
  |\{s \in S : c \in \ell(s)\}| \ge \kappa .
\end{equation}

\textbf{Coherence (maximize).}
Sampling entire documents preserves the context surrounding each segment. Beyond this, we favor samples spread across many documents rather than a few, measured by the effective number of documents:
\begin{equation}\label{eq:coherence}
  \mathcal{J}_{\mathrm{eff}}(S)
  =
  \exp\!\Bigl(-\sum_{j=1}^{|\mathcal{J}|} p_j \log p_j\Bigr),
\end{equation}
where $p_j = |S \cap J_j| / |S|$ is document $J_j$'s share of the sample. Exponentiating rescales the entropy into an interpretable document count without changing the optimum.

\textbf{Fidelity (minimize).}
We align the sample $S$ with the corpus $D$ in category composition and difficulty. The category distribution $P_C$ gives the relative frequency of each taxonomy category in $C$, and $P_e$ is a histogram of the post-edit score $e\in[0,1]$ over fixed bins, both empirical distributions computed on the corpus ($P$) and the sample ($\hat{P}$). We minimize their chi-squared divergence,
\begin{equation}\label{eq:fidelity}
  D_{\chi^2}(\hat{P}_C \,\|\, P_C)
  + \lambda\, D_{\chi^2}(\hat{P}_{e} \,\|\, P_{e}),
\end{equation}
with $\lambda>0$ trades off category against difficulty alignment.

\subsection{Method Overview}

\tactics{} formulates stratified evaluation-set construction as constrained selection over a taxonomy-induced representation of the full corpus. The framework has three stages (Figure~\ref{fig:tactics_framework}): taxonomy induction defines the category space, segment annotation maps the corpus into this space, and constrained sampling selects a fixed-budget subset that balances coverage of rare phenomena, document-level coherence, and fidelity. This yields compact evaluation sets that preserve both the linguistic structure and the key statistical properties of the full corpus.

\subsubsection{Taxonomy Extraction (Stage 1)}
\label{sec:taxonomy}

\begin{figure}[t]
    \centering
    \includegraphics[width=\linewidth, trim=5 410 300 275, clip]{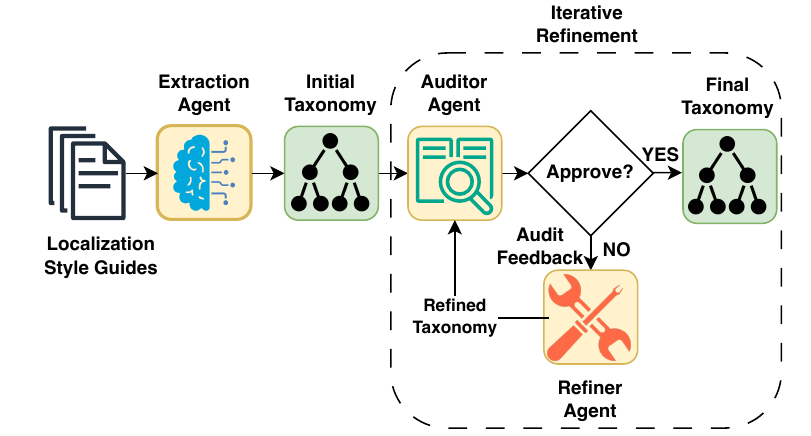}
    \caption{
    Style-guide-grounded taxonomy induction. An \emph{Extraction Agent}
    turns style guides into an initial taxonomy; an \emph{Auditor Agent}
    then checks its coverage, looping through a \emph{Refiner Agent} until
    it approves or an iteration limit is reached.
    }
    \label{fig:taxonomy_pipeline}
\end{figure}

We induce a hierarchical taxonomy $T$ from locale-specific style guides to
define the category space $C$ used for sampling. This is necessary because
the corpus does not explicitly annotate linguistic phenomena, and
embeddings or metadata alone do not capture prescriptive constraints such
as terminology, branding, formatting, or locale-specific conventions. The
taxonomy is partitioned into \textit{Deterministic} categories $C_D$,
covering high-precision rule-matchable phenomena, and \textit{Semantic}
categories $C_S$, covering context-dependent phenomena such as tone,
register, or implicit meaning.

We construct $T$ with an agentic extract--audit--refine loop
(Figure~\ref{fig:taxonomy_pipeline}; Appendix~\ref{app:taxonomy_prompts}). An Extraction Agent identifies
candidate rules from the style guide and organizes them into an initial hierarchy, an Auditor Agent checks coverage against the source, and a
Refiner Agent incorporates missing or under-specified rules. The loop
terminates when no major gaps remain or an iteration limit is reached,
yielding the prescriptive category space used for annotation, coverage
constraints, and distributional alignment in later stages.

\subsubsection{Hybrid Classification (Stage 2)}

Given taxonomy $T$, we define a labeling function $\ell(s)\subseteq C$ that maps each segment $s\in D$ to one or more taxonomy categories, embedding the corpus into the category space from Stage~1. Classification is hybrid. Deterministic categories $C_D$ are assigned by rule-based matchers, such as regular expressions for terminology or formatting patterns, producing high-precision labels with confidence $1.0$. Semantic categories $C_S$ are assigned by a context-aware classifier whose outputs are constrained to valid taxonomy paths. Each segment receives a primary category and optional secondary categories to capture co-occurring phenomena. The resulting labeled dataset $(D,\ell)$ supports the coverage constraint and defines the empirical distributions used by the fidelity objective.

\subsubsection{Stratified Sampling (Stage 3)}

Given labeled data $(D,\ell)$, we greedily select a budgeted subset
$S\subset D$, $|S|\le B$, to approximate the constrained optimization
problem of Section~\ref{sec:problem_formulation}. At each step, we add the segment that most improves the objective while preserving satisfied constraints.

\textbf{Level 1: Coverage enforcement.}
We first build a seed set that satisfies the minimum coverage constraint (Eq.~\ref{eq:coverage}) for each feasible category $c \in C$. Categories are processed from rare to frequent, and segments
are selected using a fixed mix of confidence levels. Each selected segment is augmented with a local context window from the same document to preserve translation context. This stage ensures explicit coverage of long-tail categories and initializes $S$ for later optimization.

\textbf{Level 2: Coherence optimization.}
We next improve coherence by adding documents that contribute diverse
taxonomy signals. Directly maximizing $\mathcal{J}_{\mathrm{eff}}(S)$
(Eq.~\ref{eq:coherence}) is hard, since each document's contribution
depends on its size and the evolving sample. We instead use a
category-aware document score analogous to TF--IDF, with taxonomy
categories playing the role of terms. For category $c$ and document $j$,
\begin{align}
\mathrm{CF}(c,j) &= \frac{\#\{s \in J_j : c \in \ell(s)\}}{|J_j|}, \\
\mathrm{IDF}(c) &= \log \frac{|\mathcal{J}|}{\mathrm{df}(c)},
\end{align}
where $\mathrm{df}(c)$ is the number of documents containing $c$, CF is the
category's frequency within a document, and IDF downweights categories
common across documents. To stop a category repeated within one
document from dominating, we saturate CF with hyperparameter $k_1$,
\begin{equation}
\mathrm{CF}_{\mathrm{sat}}(c,j)
=
\frac{\mathrm{CF}(c,j)(k_1+1)}{\mathrm{CF}(c,j)+k_1},
\end{equation}
and score each document as
\begin{equation}\label{eq:coh_score}
\begin{split}
\mathrm{score}(j) = {}& w_{\mathrm{conf}}(j)\, w_{\mathrm{size}}(j) \\
&\times \sum_{c \in C} \mathrm{CF}_{\mathrm{sat}}(c,j)\, \mathrm{IDF}(c),
\end{split}
\end{equation}
where $w_{\mathrm{conf}}$ favors reliable assignments and $w_{\mathrm{size}}$
downweights documents that are too small or too large
(Appendix~\ref{app:l2_ablation}, \ref{app:l2_weights}). At each step we add
the highest-scoring document's remaining segments to $S$.

\textbf{Level 3: Fidelity alignment.}
With the remaining budget, we reduce the fidelity objective
(Eq.~\ref{eq:fidelity}) to align the sample with the corpus in category
distribution and edit-rate profile. Estimating target statistics from $D$,
we add segments that close the largest gaps, selecting edited segments to
match the target difficulty profile and unedited segments from
underrepresented categories. This corrects the distributional drift
introduced by coverage and coherence optimization, yielding a sample that
faithfully reflects the corpus in both composition and difficulty.
\section{Experiments}

\subsection{Data and Baselines}
\label{sec:data_baselines}

We evaluate on four localization translation directions,
en$\rightarrow$fr, en$\rightarrow$hi, ja$\rightarrow$en, and
ja$\rightarrow$zh. Each direction uses a proprietary, in-house localization post-editing corpus in
which source segments are machine-translated and then post-edited by professional
linguists. Each segment thus carries a source, an MT translation, and a
post-edited target, and is linked to its source document. Its post-edit score is
the edit rate between the MT output and the post-edit, approximated by TER
(Section~\ref{sec:problem_formulation}). Table~\ref{tab:data_locales} summarizes the pools,
which range from 57K--84K segments, 9K--19K documents, 80--149 taxonomy
categories, and edit rates from 53.1\% to 68.0\%. We also use target-locale style guides, which specify orthographic, lexical,
and formatting constraints not observable from the corpus, to induce the
taxonomy (Section~\ref{sec:taxonomy}, example in Appendix~\ref{app:style_guide}).

All methods sample $B=5{,}000$ segments per locale from the same classified
evaluation pool. We compare \tactics{} against Random uniform sampling,
\dsir{}~\citep{xie2023dataselectionlanguagemodels} (importance resampling on
n-gram statistics), and \clusterclip{}~\citep{shao2024balanceddatasamplinglanguage}
(embedding clustering with balanced sampling).

\begin{table}[t]
\centering
\small
\resizebox{\columnwidth}{!}{%
\begin{tabular}{lcccc}
\toprule
Pair & Segments & Docs & Categories & Edit rate (\%) \\
\midrule
en $\rightarrow$ fr & 57,524 & 13,211 & 149 & 56.6 \\
en $\rightarrow$ hi & 64,932 & 18,897 & 80 & 53.1 \\
ja $\rightarrow$ en & 83,522 & 9,379 & 126 & 60.4 \\
ja $\rightarrow$ zh & 82,578 & 9,659 & 127 & 68.0 \\
\bottomrule
\end{tabular}}
\caption{Locale pairs used for baseline comparison analysis. Segments and documents from the classified evaluation pool; categories from the merged taxonomy; edit rate from post-edit annotations.}
\label{tab:data_locales}
\vspace{-1.25em}
\end{table}

\begin{table*}[t]
\centering
\small
\setlength{\tabcolsep}{3pt}
\renewcommand{\arraystretch}{0.92}
\begin{tabular}{llccccccc}
\toprule
Locale ($|C|$) & Method & Cov$_{20}|$Cov$_1$ & Gini & Pearson $r$ & MSE$_{C}\downarrow$ & MSE$_{\mathrm{TER}}\downarrow$ & Edit (\%) & S/D \\
\midrule

\multirow{4}{*}{en$\rightarrow$fr (149)}
& Random & 56|106 & 0.421 & \textbf{0.999} & 0.43 & 5.26 & 56.7 & 2.0 \\
& \dsir{} & 13|70 & 0.133 & 0.983 & \textbf{0.38} & \textbf{1.46} & 64.8 & 1.2 \\
& \clusterclip{} & 56|101 & 0.426 & 0.999 & 1.23 & 24.6 & 58.2 & 2.0 \\
& \textbf{\tactics{}} & \textbf{70|133} & \textbf{0.497} & 0.982 & 6.90 & 95.9 & \textbf{56.6} & \textbf{3.1} \\

\midrule
\multirow{4}{*}{en$\rightarrow$hi (80)}
& Random & 64|79 & 0.388 & \textbf{0.999} & 3.66 & \textbf{7.98} & 53.4 & 1.8 \\
& \dsir{} & 9|70 & 0.164 & 0.952 & 49.6 & 66.4 & 56.8 & 1.2 \\
& \clusterclip{} & 66|80 & 0.393 & 0.999 & \textbf{1.24} & 14.9 & 55.5 & 1.8 \\
& \textbf{\tactics{}} & \textbf{79|80} & \textbf{0.468} & 0.969 & 46.5 & 133 & \textbf{53.1} & \textbf{2.4} \\

\midrule
\multirow{4}{*}{ja$\rightarrow$en (126)}
& Random & 53|85 & 0.349 & \textbf{1.000} & \textbf{0.09} & \textbf{11.5} & 59.9 & 1.9 \\
& \dsir{} & 11|57 & 0.151 & 0.892 & 49.8 & 843 & 81.0 & 1.2 \\
& \clusterclip{} & 54|87 & 0.331 & 0.942 & 26.8 & 1090 & 74.0 & 1.8 \\
& \textbf{\tactics{}} & \textbf{85|102} & \textbf{0.491} & 0.972 & 11.8 & 111 & \textbf{60.4} & \textbf{3.4} \\

\midrule
\multirow{4}{*}{ja$\rightarrow$zh (127)}
& Random & 52|96 & 0.346 & \textbf{0.999} & \textbf{0.08} & \textbf{6.02} & 67.8 & 1.9 \\
& \dsir{} & 7|57 & 0.114 & 0.893 & 50.2 & 1510 & 88.1 & 1.1 \\
& \clusterclip{} & 51|94 & 0.308 & 0.929 & 25.1 & 1690 & 85.7 & 1.7 \\
& \textbf{\tactics{}} & \textbf{59|113} & \textbf{0.550} & 0.949 & 9.17 & 71.6 & \textbf{68.0} & \textbf{4.0} \\

\bottomrule
\end{tabular}
\caption{Intrinsic sampling quality at $B=5{,}000$. Cov$_{20}|$Cov$_1$ counts
categories with $\geq20$ and $\geq1$ samples; $r$ is Pearson correlation with
the full-corpus category distribution; MSE$_{C}$ and MSE$_{\mathrm{TER}}$
($\times10^{-6}$) are mean squared errors of the sampled vs.\ full-corpus
category and TER distributions; Edit is post-edit rate; S/D is segments per
document. Random matches the corpus distributions most closely, while
\tactics{} trades fidelity to cover rare categories and harder segments.
Baselines: \dsir{}~\citep{xie2023dataselectionlanguagemodels},
\clusterclip{}~\citep{shao2024balanceddatasamplinglanguage}.}
\label{tab:intrinsic_main}
\end{table*}

\begin{table*}[t]
\centering
\scriptsize
\setlength{\tabcolsep}{3pt}
\renewcommand{\arraystretch}{0.92}
\begin{tabular}{ll cccc cccc}
\toprule
& & \multicolumn{4}{c}{\awstranslate{}} & \multicolumn{4}{c}{\llmmt{}} \\
\cmidrule(lr){3-6} \cmidrule(lr){7-10}
Locale & Metric & Rand. & \clusterclip{} & \dsir{} & \tactics{} 
             & Rand. & \clusterclip{} & \dsir{} & \tactics{} \\
\midrule

\multirow{4}{*}{en$\rightarrow$fr}
& TER$\downarrow$  & 0.309 & \textbf{0.305} & 0.309 & 0.315 & 0.267 & 0.252 & 0.257 & \textbf{0.231} \\
& BLEU$\uparrow$ & 0.655 & 0.662 & 0.655 & \textbf{0.668} & 0.686 & 0.699 & 0.690 & \textbf{0.717} \\
& chrF$\uparrow$ & 0.795 & 0.800 & 0.798 & \textbf{0.804} & 0.799 & 0.812 & 0.807 & \textbf{0.833} \\
& WH$\uparrow$   & \textbf{6.50} & 5.48 & 6.24 & 5.64 & 7.80 & 5.90 & 7.18 & \textbf{11.84} \\

\midrule
\multirow{4}{*}{en$\rightarrow$hi}
& TER$\downarrow$  & 0.349 & 0.357 & 0.409 & \textbf{0.343} & 0.322 & 0.328 & 0.341 & \textbf{0.320} \\
& BLEU$\uparrow$ & 0.600 & 0.606 & 0.608 & \textbf{0.625} & 0.636 & 0.633 & 0.612 & \textbf{0.641} \\
& chrF$\uparrow$ & 0.716 & 0.718 & 0.700 & \textbf{0.730} & 0.699 & 0.702 & 0.686 & \textbf{0.760} \\
& WH$\uparrow$   & 1.55 & 1.70 & 1.45 & \textbf{2.50} & 3.40 & 3.65 & 3.20 & \textbf{15.38} \\

\midrule
\multirow{4}{*}{ja$\rightarrow$en}
& TER$\downarrow$  & \textbf{0.410} & 0.422 & 0.422 & 0.560 & 0.323 & 0.301 & \textbf{0.299} & 0.328 \\
& BLEU$\uparrow$ & 0.619 & 0.616 & \textbf{0.639} & 0.634 & 0.728 & 0.753 & 0.750 & \textbf{0.795} \\
& chrF$\uparrow$ & 0.718 & 0.710 & 0.737 & \textbf{0.745} & 0.808 & 0.827 & 0.821 & \textbf{0.847} \\
& WH$\uparrow$   & 5.25 & \textbf{6.39} & 6.00 & 4.93 & 6.34 & 6.78 & 6.12 & \textbf{31.79} \\

\midrule
\multirow{4}{*}{ja$\rightarrow$zh}
& TER$\downarrow$  & 0.306 & 0.319 & \textbf{0.302} & 0.319 & 0.146 & 0.116 & 0.145 & \textbf{0.090} \\
& BLEU$\uparrow$ & 0.576 & 0.561 & 0.585 & \textbf{0.590} & 0.788 & 0.833 & 0.790 & \textbf{0.865} \\
& chrF$\uparrow$ & 0.613 & 0.610 & 0.620 & \textbf{0.713} & 0.814 & 0.854 & 0.815 & \textbf{0.907} \\
& WH$\uparrow$   & 4.75 & 4.60 & \textbf{4.90} & 3.98 & 5.85 & 5.70 & 5.90 & \textbf{27.38} \\

\bottomrule
\end{tabular}
\caption{Downstream translation performance across sampling strategies.
Best value per direction, metric, and system is in \textbf{bold}. Lower TER
is better, and higher BLEU, chrF, and WH are better. WH is the withholding
rate, the percentage of segments passing automated QA checks. Baselines
include \dsir{}~\citep{xie2023dataselectionlanguagemodels} and
\clusterclip{}~\citep{shao2024balanceddatasamplinglanguage}.}
\label{tab:downstream_sampling}
\end{table*}

\subsection{Evaluation Setup}
\label{sec:eval_setup}

We select pipeline models by role, as the method is agnostic to the specific
LLMs. Taxonomy induction runs once per locale, so we use \textbf{Claude Sonnet 4.5}
to generate the taxonomy and the more capable \textbf{Claude Opus 4.5} to audit
it for missing or under-specified rules
(Figure~\ref{fig:taxonomy_pipeline}, Section~\ref{sec:taxonomy}). Classification
runs over the whole corpus, so we label segments with \textbf{Kimi K2.5},
constraining its outputs to valid taxonomy paths. From a different family than
the taxonomy models, it labels cheaply at scale and independently of the
generator. For downstream evaluation we compare
\awstranslate{}, a dedicated neural MT system, with \llmmt{}, a
\textbf{Gemma 4-26B-A4B} model prompted to translate, conditioned on summarized
style guides.

We report intrinsic metrics aligned with the three objectives of \tactics{}.
\textbf{Coverage} is Cov$_{20}$ and Cov$_1$, the number of categories with at
least 20 and at least 1 sampled segments. \textbf{Coherence} combines the
Gini coefficient over per-document segment counts with segments per document
(S/D), where higher values mean more multi-segment context. \textbf{Fidelity} compares the sample to the full corpus
via Pearson correlation ($r$) and category- and TER-distribution MSE (higher
$r$, lower MSE better), with edit rate as a difficulty proxy. Downstream, we
report TER~\citep{snover2006study}, BLEU~\citep{bleu2002}, and
chrF~\citep{popovic-2015-chrf} against the human post-edited target, together
with withholding rate (WH), the fraction of segments passing an automated QA pipeline.
The pipeline applies deterministic checks (a translation-memory match against
known-good references, and length heuristics that flag implausible
target-to-source length ratios or name lengths) together with a learned
classifier that predicts translation quality from the source and target sentence.
These checks use no style guides, so they are independent of the taxonomy and the
\llmmt{} prompt and favor neither system.

\subsection{Results}
\label{sec:results}
We provide taxonomy consistency checks, including inter-model agreement and
manual spot checks, in Appendix~\ref{app:taxonomy_validation}.

\paragraph{Intrinsic sampling quality (Table~\ref{tab:intrinsic_main}).}
\tactics{} improves coverage and coherence across all directions.
\covtwenty{} rises over Random from 56 to 70, 64 to 79, 53 to 85, and 52 to
59 for en$\rightarrow$fr, en$\rightarrow$hi, ja$\rightarrow$en, and
ja$\rightarrow$zh. It also attains the highest Gini and segments per document
(S/D) everywhere, selecting multiple segments from the same document rather
than isolated ones. This matters because translation quality is
context-dependent, and a segment like ``Click here'' is only translatable
alongside its neighbors. But covering the tail pulls the sample away from the corpus. Random most
closely matches the full-corpus category and TER distributions (lowest MSE),
since it simply reproduces the corpus proportions, whereas \tactics{}
deviates more to represent rare categories and harder segments. Even so,
\tactics{} keeps the corpus category ordering (Pearson $r=0.949$--$0.982$)
and stays far more stable than \dsir{} and \clusterclip{}, whose MSE swings
by orders of magnitude across directions.

\begin{table}[t]
\centering
\small
\setlength{\tabcolsep}{4pt}
\renewcommand{\arraystretch}{0.95}
\begin{tabular}{llrrl}
\toprule
Locale & Systems & Segs & True $\Delta$ & Verdict \\
\midrule
en$\rightarrow$fr & NMT / LLM & 3959 & $+0.016$ & not sig. \\
en$\rightarrow$hi & NMT / LLM & 4334 & $+0.314$ & significant \\
ja$\rightarrow$en & LLM$^{-\mathrm{SG}}$ / NMT & 3168 & $+1.121$ & significant \\
ja$\rightarrow$zh & LLM / NMT & 3062 & $-0.086$ & significant \\
\bottomrule
\end{tabular}
\caption{Full-pool human-audit reference verdicts, the ground truth each
subsample aims to reproduce. Systems: NMT (\awstranslate{}), LLM (\llmmt{}, style-guided),
and LLM$^{-\mathrm{SG}}$ (\llmmt{} without style guides);
ja$\rightarrow$en uses LLM$^{-\mathrm{SG}}$, the only direction it was
annotated. Each segment is judged by one linguist with a
second-expert QA pass. MQM penalty $=25\,\text{critical}+5\,\text{major}
+\text{minor}$ (lower is better), so $\Delta=A-B>0$ means the first-listed
system $A$ is worse. Full $t$ and $p$ values in Appendix~\ref{app:estimation}.}
\label{tab:reference_verdicts}
\end{table}

\paragraph{Downstream translation performance (Table~\ref{tab:downstream_sampling}).}
On a random sample, \awstranslate{} and \llmmt{} score very close across TER,
BLEU, chrF, and WH, since it rarely includes the segments that separate them.
The \tactics{} set covers these rarer and harder cases, and the gap becomes
clear. On ja$\rightarrow$zh the
\llmmt{}--\awstranslate{} WH gap widens from about one point to over twenty,
BLEU from 0.788 to 0.865 and chrF from 0.814 to 0.907, and the mean WH gap
across directions grows from 1.3 to 17.3 points. Human evaluation shows the
same pattern by category. Auditing one NMT system on comparably sized
\tactics{} and random samples ($\sim$3{,}000 vs.\ 2{,}874 segments), the
\tactics{} audit covered 18 error categories against 15 for random, including
three the random audit missed entirely, \texttt{number-format},
\texttt{text-truncation}, and \texttt{offensive-content}. Several of these
carry major errors, three for text-truncation and one for offensive-content,
so a random-only audit leaves real, category-specific failures unmeasured. We
therefore use the two together, \tactics{} to check per-category quality
including the long tail, and random to estimate overall quality at the corpus
proportion.

\begin{figure*}[t]
    \centering
    \includegraphics[width=\textwidth]{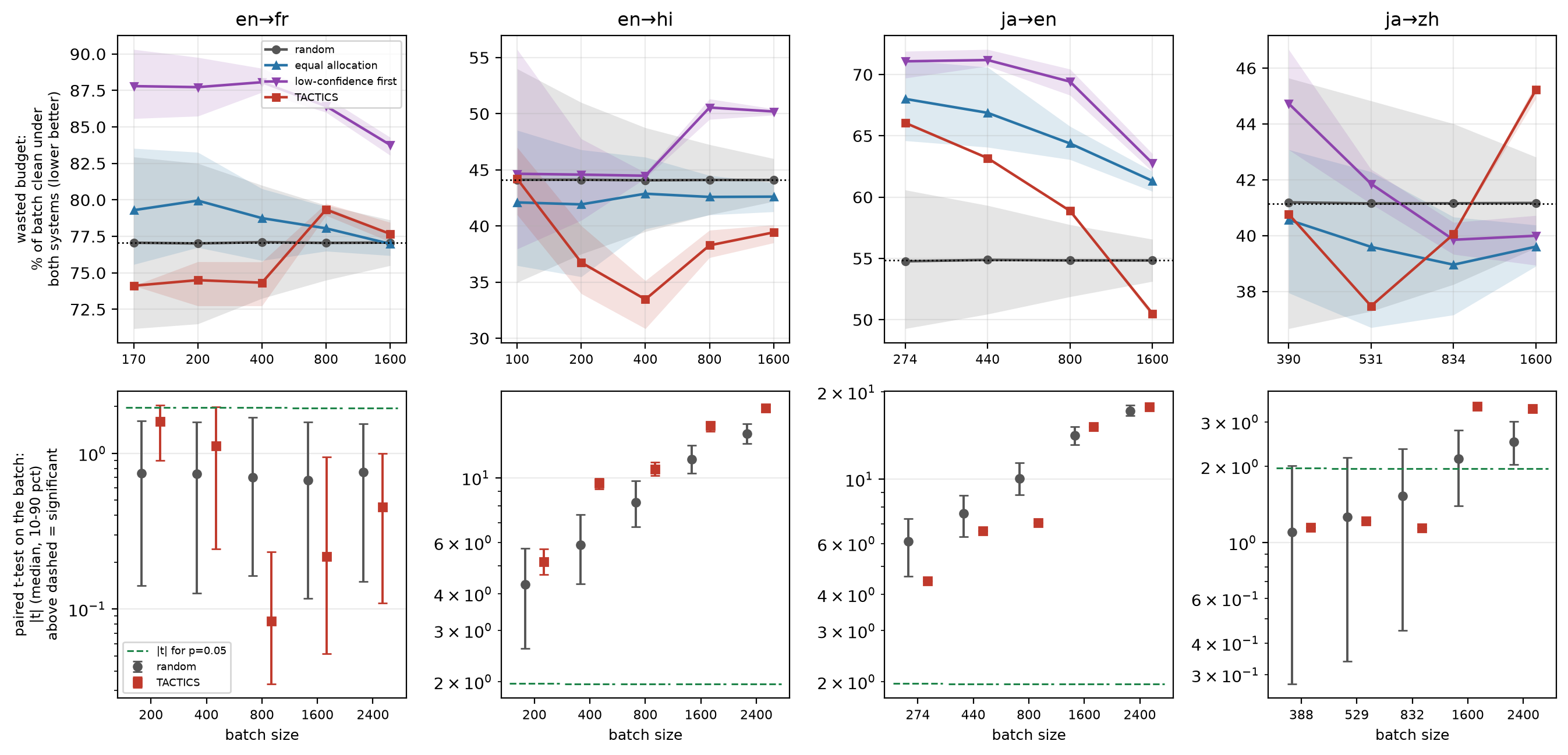}
    \caption{Sample efficiency across four directions. \textbf{Top}, wasted
    budget, the percentage of a batch that is clean under both systems and so
    cannot separate them (lower is better). \textbf{Bottom}, the paired
    $|t|$ on the batch (median over draws, 10--90 percentile bars). Points
    above the dashed line are significant at $p{=}0.05$. \tactics{} is
    compared against random, equal-allocation, and low-confidence-first
    selection.}
    \label{fig:sample_efficiency}
\end{figure*}

\paragraph{Sample efficiency (Table~\ref{tab:reference_verdicts}, Figure~\ref{fig:sample_efficiency}).}
Table~\ref{tab:reference_verdicts} gives the true system ranking each
subsample must recover, and Figure~\ref{fig:sample_efficiency} measures how
efficiently the samplers reach it. The four selectors are Random (uniform,
unstratified), Equal-allocation (flat per-category quota), Low-confidence-first
(lowest classifier-confidence segments, \tactics{}'s hard-example heuristic
alone), and \tactics{} (the full three-level pipeline). The three category-aware
selectors differ only in how they spend the budget, isolating \tactics{}'s
allocation strategy rather than the corpus-level baselines of
Section~\ref{sec:data_baselines}. The top row is wasted budget, the fraction
of a batch that both systems translate cleanly and that reveals nothing about
their difference, which \tactics{} keeps lowest in most settings. The bottom
row is the paired $t$-statistic magnitude $|t|$ per batch, larger when a batch
separates the systems more clearly and significant above the dashed line. On
en$\rightarrow$hi \tactics{} exceeds random at every budget (ratio
$1.19$--$1.63$), and on ja$\rightarrow$zh it clears the threshold at 1600 and
2400 segments where random does not. en$\rightarrow$fr is a control, where the
systems are tied ($\Delta=0.016$) and both samplers stay below the line, so
\tactics{} signals a difference only when one is real. Although it estimates the corpus average less accurately
(Appendix~\ref{app:estimation}), \tactics{} detects a real gap from far fewer
segments than random.

\section{Discussion}
\label{sec:discussion}

\paragraph{A representative sample is not a diagnostic one.}
Random sampling reproduces the corpus almost exactly, which is why it
struggles to compare systems. Most of a corpus is routine text both systems
handle, so a faithful sample is dominated by segments that carry no signal
about their difference, while the cases that separate systems, such as rare
terminology, date and number formats, or long-context documents, are easily
left out. A sample can thus match the corpus yet fail to reveal which system
is better.

\paragraph{Generic selectors miss prescriptive rules.}
\dsir{} and \clusterclip{} rank segments by lexical or embedding similarity,
capturing frequent surface patterns but not the prescriptive rules
localization depends on, such as terminology, punctuation, and date, currency,
and locale-specific formats. They therefore cover fewer taxonomy categories
than \tactics{}, and their sample statistics vary widely across directions.

\paragraph{Coverage trades against fidelity by design.}
By guaranteeing rare categories and harder segments, \tactics{} deviates from
the corpus more than random and so has higher distributional error. This is
intentional. \tactics{} keeps the corpus category ordering while shifting
weight to the long tail, making it a weaker estimator of the corpus average
but a stronger detector of system differences. Fidelity is a constraint to
stay close enough, not the objective to maximize.

\paragraph{Use both samplers together.}
The two roles are complementary. Human evaluation is expensive, so the
limited annotation budget is best spent on a \tactics{} set, which guarantees
that every content type is judged, including the rare and difficult cases
random evaluation can miss. A cheaper random sample then estimates overall
quality at the corpus proportion. Together they answer whether a system is
good everywhere and how it performs on average.
\section{Conclusion and Future Work}
\label{sec:conclusion}

We presented \tactics{}, a taxonomy-aware method for building compact,
diagnostic MT evaluation sets. By inducing categories from localization style
guides and jointly optimizing coverage of rare phenomena, document-level
coherence, and distributional fidelity, \tactics{} guarantees that evaluation
covers the content types that distinguish systems, including the long tail
that random sampling misses. Across four directions it improves category
coverage over random, lexical, and embedding-based selection, and recovers
the true system ranking from fewer segments where a real quality gap exists.
It does so by trading corpus-average fidelity for discriminative power, so we
recommend pairing a \tactics{} set, where costly human evaluation is best
spent, with a random sample that estimates overall quality at the corpus
proportion.

The core mechanism, taxonomy-aware constrained sampling over coverage,
coherence, and fidelity, is not specific to machine translation. Any
evaluation setting with a prescriptive taxonomy, long-tail skew over that
taxonomy, and a fixed annotation budget could benefit. Future work includes
validating \tactics{} on public benchmarks and style guides to support
reproducibility, extending it to other domains such as content moderation and
clinical text, expanding human validation of taxonomy boundaries, and
studying adaptive sampling budgets.

\section*{Limitations}

Our primary evaluation uses a proprietary localization corpus that we cannot release, which limits direct reproducibility. We therefore plan a public companion study on open style guides and benchmarks. The evaluation is also confined to localization, and generalization to other domains such as news, biomedical, or legal text requires further validation. The taxonomy induction pipeline depends on the availability and quality of prescriptive style guides, which may not exist or may vary in granularity across locales, and the LLM-based classifier reaches only moderate inter-annotator agreement (Cohen's $\kappa = 0.51$), so misclassifications can affect coverage. By design, \tactics{} trades fidelity to the corpus distribution for coverage and discriminative power, making it a deliberately biased estimator of corpus-average quality that should be paired with a random sample when an aggregate estimate is needed.


\section*{Acknowledgments}

We thank Elijah Smith for automating and
containerizing the sampling pipeline, enabling large-scale sampling jobs to be
scheduled and run reliably with minimal manual intervention. This
infrastructure was instrumental in carrying out the experiments presented in
this paper.

\bibliography{custom}

\appendix

\appendix
\section*{Appendix}
\addcontentsline{toc}{section}{Appendix}

\section{Level-2 Scoring Ablation}
\label{app:l2_ablation}

Level-2 scores each document by how much rare-category evidence it
concentrates (Eq.~\eqref{eq:coh_score}). Table~\ref{tab:l2_ablation}
ablates the two design choices in that function, \emph{saturation} and
\emph{aggregation}, under a fixed budget $B=5{,}000$ (seed 42) with Level-1
disabled, so the differences reflect Level-2 scoring alone. The score
combines two per-category signals. \emph{Category frequency}
$\mathrm{CF}(c,j)$ is the fraction of segments in document $j$ assigned to
category $c$ and is high when the document gives concentrated evidence for a
phenomenon. \emph{Inverse document frequency} $\mathrm{IDF}(c)$ is large for
rare categories, so the product $\mathrm{CF}\cdot\mathrm{IDF}$ favors
documents rich in rare phenomena rather than documents that only repeat
common ones. \emph{Saturation} applies a concave transform
$\mathrm{CF}_{\mathrm{sat}}$ so that a category repeated many times within
one document cannot dominate its score. \emph{Aggregation} is either
\emph{pure}, where each category contributes once as
$\mathrm{CF}_{\mathrm{sat}}(c,j)\,\mathrm{IDF}(c)$ (Eq.~\eqref{eq:coh_score}),
or \emph{weighted}, which additionally scales each term by the segment count
$n(c,j)=|\{s \in J_j : c \in \ell(s)\}|$ and rewards repeated occurrences.
All four variants share the same confidence weight
$w_{\mathrm{conf}} \in \{1.2,1.0,0.8\}$ (high, medium, low) and size factor
$w_{\mathrm{size}}$ (Appendix~\ref{app:l2_weights}). The table shows a sharp
split between the two aggregation choices. Weighted variants collapse the
budget into 16--17 documents (Gini $\approx 0.1$, Top-10 $\approx 66$--$69\%$)
and cover only 37/76 categories. Pure variants instead spread selection over
hundreds to more than 1{,}600 documents, raising effective diversity by more
than an order of magnitude and lifting coverage to 48--49/76.
\textbf{Saturated pure} trades a small amount of raw diversity for more
balanced within-document evidence, selecting 815 documents at Gini $0.441$
and Cov$_{30}=48/76$, and is our default Level-2 configuration.

\begin{table}[t]
\centering
\footnotesize
\setlength{\tabcolsep}{2.5pt}
\resizebox{\columnwidth}{!}{%
\begin{tabular}{lrrcrrr}
\toprule
Config & Docs & Segs & Gini & Eff.~D & Top-10 & Cov$_{30}$ \\
\midrule
CF--IDF (wt.) & 17 & 5,194 & 0.113 & 16.6 & 66.3\% & 37/76 \\
CF--IDF (pure) & 1,622 & 5,150 & \textbf{0.491} & \textbf{959.6} & \textbf{8.0\%} & \textbf{49/76} \\
CF--IDF (sat., wt.) & 16 & 5,032 & 0.094 & 15.8 & 68.5\% & 37/76 \\
\textbf{CF--IDF (sat., pure)} & \textbf{815} & \textbf{5,007} & 0.441 & 551.3 & 9.2\% & 48/76 \\
\bottomrule
\end{tabular}}
\caption{Level-2 scoring ablation under fixed budget ($B=5{,}000$) with
Level-1 disabled. Cov$_{30}$: categories with $\geq$30 segments.}
\label{tab:l2_ablation}
\end{table}

\section{Level-2 Weighting Details}
\label{app:l2_weights}

The Level-2 document score (Eq.~\ref{eq:coh_score}) applies two
multiplicative weights that favor reliable, moderately sized documents. The
confidence weight $w_{\mathrm{conf}}(j)$ is $1.2$ for high-confidence
documents, $1.0$ for medium, and $0.8$ for low. For the size weight, let
$n_j = |J_j|$ be the number of segments in document $j$, let $Q_1$ be the
first quartile of document sizes, and let $n_{\max}$ be a segment cap, and
define $r_j = \min\!\bigl(\frac{n_j - n_{\max}}{n_{\max}},\, 1\bigr)$. The
size weight is
\begin{equation}
\small
w_{\mathrm{size}}(j) \!=\!
\begin{cases}
0.8 + 0.2 \cdot \frac{n_j}{Q_1}, & n_j \leq Q_1, \\[3pt]
1.0, & Q_1 < n_j \leq n_{\max}, \\[3pt]
\max\bigl(0.5,\, 1 - 0.3\, r_j\bigr), & n_j > n_{\max}.
\end{cases}
\end{equation}
This down-weights both highly fragmented samples drawn from very small
documents and budget domination by very large ones.

\section{Example Style Guide Snippet}
\label{app:style_guide}

\begin{tcolorbox}[
  colback=gray!5,
  colframe=gray!60,
  title=\textbf{Style Guide Snippet (Illustrative)},
  fonttitle=\bfseries,
  boxrule=0.5pt,
  arc=2pt
]
\small

\textbf{Currency Format.}
The currency symbol should be inserted after the number and preceded by a
space. For Canadian dollars, there should be a space between the dollar
sign and the currency symbol.

\textbf{Examples:} 50,99~€ / 4~500~EUR / 50~\$~CA

As a general rule, use ISO currency codes.

\medskip

\textbf{Common currencies:}

\resizebox{\linewidth}{!}{%
\begin{tabular}{llll}
\toprule
Country/Region & Currency name & Symbol & ISO code \\
\midrule
EU & euro & € & EUR \\
Canada & Canadian dollar & CA \$ & CAD \\
USA & US dollar & \$ & USD \\
\bottomrule
\end{tabular}}

\medskip

If currency symbols or ISO codes (e.g., \$, €, GBP, EUR) appear in the
source text, they should be preserved and formatted according to
locale-specific rules.

\medskip

\textbf{Date Format.}
Dates must follow the French day--month format rather than the US
month--day format.

\medskip

\resizebox{\linewidth}{!}{%
\begin{tabular}{lll}
\toprule
US English & French & Comments \\
\midrule
7/14/2013 & 14/07/2013 & N/A \\
July 14, 2013 & 14 juillet 2013 & Extended \\
\bottomrule
\end{tabular}}

\end{tcolorbox}

\noindent
This snippet is a representative portion of a larger localization style
guide. It illustrates the type of constraints used to induce the taxonomy in
Section~\ref{sec:taxonomy}.

\section{Per-Level Pipeline Breakdown}
\label{app:per_level_breakdown}

Each level of \tactics{} contributes a distinct sampling property. Table~\ref{tab:per_level_breakdown}
reports per-stage statistics to illustrate how coverage, coherence, and
fidelity are introduced incrementally. L1 constructs the initial
coverage-oriented seed set, L2 adds coherent document-level context, and L3
fills the remaining budget to match the full-corpus edit-rate statistics.

\begin{table*}[t]
\centering
\small
\setlength{\tabcolsep}{4pt}
\renewcommand{\arraystretch}{0.95}
\begin{tabular}{llr l r c c r}
\toprule
Locale & Dir & $|C|$ & Stage & Seg. & Cov$_{20}$ & Edit (\%) & L2 Docs \\
\midrule

\multirow{3}{*}{fr-FR} & \multirow{3}{*}{en$\rightarrow$fr} & \multirow{3}{*}{105}
& L1 & 2,882 & -- & -- & -- \\
& & & L1+L2 & 3,503 & 98/105 & 47.6 & 169 \\
& & & L1+L2+L3 & 5,000 & 98/105 & \textbf{45.7} & 169 \\

\midrule
\multirow{3}{*}{ja-JP} & \multirow{3}{*}{en$\rightarrow$ja} & \multirow{3}{*}{76}
& L1 & 1,897 & -- & -- & -- \\
& & & L1+L2 & 3,524 & 75/76 & 75.7 & 302 \\
& & & L1+L2+L3 & 5,000 & 75/76 & \textbf{61.5} & 302 \\

\midrule
\multirow{3}{*}{ko-KR} & \multirow{3}{*}{en$\rightarrow$ko} & \multirow{3}{*}{86}
& L1 & 2,255 & -- & -- & -- \\
& & & L1+L2 & 3,504 & 83/86 & 68.2 & 331 \\
& & & L1+L2+L3 & 5,000 & 83/86 & \textbf{55.4} & 331 \\

\midrule
\multirow{3}{*}{hi-IN} & \multirow{3}{*}{en$\rightarrow$hi} & \multirow{3}{*}{80}
& L1 & 1,510 & -- & -- & -- \\
& & & L1+L2 & 3,506 & 79/80 & 74.2 & 483 \\
& & & L1+L2+L3 & 5,000 & 79/80 & \textbf{53.1} & 483 \\

\midrule
\multirow{3}{*}{en-US} & \multirow{3}{*}{ja$\rightarrow$en} & \multirow{3}{*}{126}
& L1 & 2,913 & -- & -- & -- \\
& & & L1+L2 & 3,510 & 86/126 & 64.6 & 53 \\
& & & L1+L2+L3 & 5,000 & 86/126 & \textbf{60.4} & 53 \\

\midrule
\multirow{3}{*}{zh-CN} & \multirow{3}{*}{ja$\rightarrow$zh} & \multirow{3}{*}{127}
& L1 & 2,530 & -- & -- & -- \\
& & & L1+L2 & 3,507 & 93/127 & 73.3 & 105 \\
& & & L1+L2+L3 & 5,000 & 93/127 & \textbf{68.0} & 105 \\

\bottomrule
\end{tabular}
\caption{Per-stage pipeline statistics at $N=5{,}000$. Cov$_{20}$ denotes
the number of taxonomy categories with at least 20 sampled segments. Bold
edit rates indicate the final L3-aligned edit rate, matching the full
evaluation pool.}
\label{tab:per_level_breakdown}
\end{table*}

\paragraph{Level contributions.}
L1 selects the initial coverage-oriented seed set, ranging from roughly
1.5K to 2.9K segments across locales. L2 then adds coherent document-level
context to reach approximately 3.5K segments. At this point coverage is
effectively fixed, for example fr-FR reaches 98/105 categories and hi-IN
reaches 79/80. L2 also introduces coherent multi-segment documents, with the
number of selected documents varying by locale and document-size
distribution.

L3 uses the remaining budget to reach $N=5{,}000$ and align the sample
with the full-corpus edit-rate profile. Since L3 only adds segments,
coverage does not decrease from L2 to L3. The edit-rate correction can be
substantial, with ja-JP decreasing from 75.7\% to 61.5\% and hi-IN from
74.2\% to 53.1\%. This happens because L1 and L2 over-sample edited
segments, which are more category-diverse, while L3 rebalances the final
sample by adding segments that restore distributional fidelity.

\section{Cross-Locale Generalization}
\label{app:cross_locale}

We evaluate \tactics{} across 22 locales spanning seven language families
to assess whether the sampling objectives generalize beyond the main
baseline-comparison directions. Table~\ref{tab:cross_locale} reports
intrinsic sampling quality for the test split at $N=5{,}000$. Across
locales, \tactics{} maintains strong distributional fidelity, with mean
Pearson correlation $r=0.992$ and mean edit-rate deviation of 0.5
percentage points. L1 coverage varies with taxonomy size, where compact
taxonomies such as en-AE (48 categories) achieve higher coverage at the
$\geq20$ threshold, while larger taxonomies such as en-AU (177 categories)
are harder to cover under the same fixed budget. L2 Gini averages 0.54
across locales, indicating consistent document-level diversity.

\begin{table}[t]
\centering
\small
\setlength{\tabcolsep}{4pt}
\renewcommand{\arraystretch}{0.92}
\resizebox{\columnwidth}{!}{%
\begin{tabular}{llrrrr}
\toprule
Locale & Family & $|C|$ & Cov$_{20}$ & Gini & $r$ / Edit $\Delta$ \\
\midrule
es-MX & Romance & 76 & 69/76 & 0.441 & 0.980 / 0.1 \\
zh-TW & Sinitic & 87 & 48/87 & 0.490 & 0.987 / 0.0 \\
en-AE & Germanic & 48 & 35/48 & 0.576 & 0.991 / 1.7 \\
en-AU & Germanic & 177 & 36/177 & 0.626 & 0.997 / 0.3 \\
en-CA & Germanic & 82 & 38/82 & 0.571 & 0.995 / 1.5 \\
en-GB & Germanic & 95 & 32/95 & 0.747 & 0.951 / 0.0 \\
en-IE & Germanic & 128 & 34/128 & 0.610 & 0.994 / 1.9 \\
en-IN & Germanic & 112 & 26/112 & 0.584 & 0.999 / 0.6 \\
en-SG & Germanic & 72 & 40/72 & 0.537 & 0.996 / 1.5 \\
en-US & Germanic & 126 & 34/126 & 0.491 & 0.997 / 0.4 \\
en-ZA & Germanic & 136 & 34/136 & 0.605 & 0.992 / 0.8 \\
nl-BE & Germanic & 89 & 44/89 & 0.517 & 0.994 / 0.2 \\
nl-NL & Germanic & 100 & 47/100 & 0.525 & 0.996 / 0.0 \\
hi-IN & Indic & 80 & 51/80 & 0.468 & 0.996 / 0.0 \\
ja-JP & Japonic & 76 & 48/76 & 0.526 & 0.992 / 0.0 \\
ko-KR & Koreanic & 86 & 42/86 & 0.500 & 0.997 / 0.0 \\
fr-BE & Romance & 77 & 45/77 & 0.511 & 0.994 / 0.0 \\
fr-FR & Romance & 105 & 44/105 & 0.541 & 0.996 / 0.3 \\
it-IT & Romance & 96 & 45/96 & 0.510 & 0.995 / 0.2 \\
pt-BR & Romance & 150 & 42/150 & 0.474 & 0.997 / 0.1 \\
pt-PT & Romance & 92 & 44/92 & 0.516 & 0.991 / 0.2 \\
pl-PL & Slavic & 87 & 37/87 & 0.482 & 0.992 / 0.0 \\
\midrule
\textbf{Mean} & -- & \textbf{99} & \textbf{44.0\%} & \textbf{0.543} & \textbf{0.992 / 0.5} \\
\bottomrule
\end{tabular}}
\caption{Cross-locale intrinsic sampling quality at $N=5{,}000$. Cov$_{20}$
reports categories with at least 20 samples. Edit $\Delta$ is the absolute
edit-rate deviation from the full evaluation pool, in percentage points.}
\label{tab:cross_locale}
\end{table}

\section{Corpus-Mean Estimation Error}
\label{app:estimation}

Table~\ref{tab:estimation_rmse} reports the RMSE of the estimated system
delta against the full-pool truth for random and \tactics{} batches. \tactics{}
has higher RMSE in most cells, and the error is bias-dominated
(bias$^2$/MSE near 100\%). This is expected of a deliberately unrepresentative
diagnostic batch, where concentrating rare and harder segments makes the
sample a poor estimator of the corpus average while making it a stronger
detector of the system ranking. This trade-off motivates pairing \tactics{}
with a separate random sample for aggregate estimation.

\begin{table}[t]
\centering
\small
\setlength{\tabcolsep}{5pt}
\renewcommand{\arraystretch}{0.95}
\resizebox{\columnwidth}{!}{%
\begin{tabular}{lrrrrr}
\toprule
Locale & Batch & RMSE$_{\text{rand}}$ & RMSE$_{\text{TAC}}$ & DEFF & bias$^2$/MSE \\
\midrule
\multirow{5}{*}{en$\rightarrow$fr}
 & 200  & 0.0763 & 0.1010 & 1.75  & 93\% \\
 & 400  & 0.0532 & 0.0627 & 1.39  & 79\% \\
 & 800  & 0.0355 & 0.0138 & 0.15  & 88\% \\
 & 1600 & 0.0218 & 0.0272 & 1.55  & 85\% \\
 & 2400 & 0.0144 & 0.0280 & 3.79  & 87\% \\
\midrule
\multirow{5}{*}{en$\rightarrow$hi}
 & 200  & 0.0747 & 0.0860 & 1.32  & 88\% \\
 & 400  & 0.0513 & 0.2269 & 19.61 & 99\% \\
 & 800  & 0.0346 & 0.0928 & 7.20  & 97\% \\
 & 1600 & 0.0212 & 0.0750 & 12.54 & 98\% \\
 & 2400 & 0.0148 & 0.0679 & 20.93 & 99\% \\
\midrule
\multirow{5}{*}{ja$\rightarrow$en}
 & 274  & 0.1846 & 0.6644 & 12.95 & 100\% \\
 & 440  & 0.1428 & 0.5706 & 15.97 & 100\% \\
 & 800  & 0.0947 & 0.3493 & 13.60 & 100\% \\
 & 1600 & 0.0556 & 0.4533 & 66.50 & 100\% \\
 & 2400 & 0.0320 & 0.1628 & 25.92 & 100\% \\
\midrule
\multirow{5}{*}{ja$\rightarrow$zh}
 & 388  & 0.0809 & 0.1052 & 1.69  & 100\% \\
 & 529  & 0.0667 & 0.0657 & 0.97  & 100\% \\
 & 832  & 0.0492 & 0.0082 & 0.03  & 100\% \\
 & 1600 & 0.0292 & 0.0874 & 8.97  & 100\% \\
 & 2400 & 0.0162 & 0.0375 & 5.32  & 100\% \\
\bottomrule
\end{tabular}}
\caption{Corpus-mean estimation error. RMSE of the estimated system delta
against the full-pool truth, for random and \tactics{} batches.
$\text{DEFF}=(\text{RMSE}_{\text{TAC}}/\text{RMSE}_{\text{rand}})^2$, and values
below 1 favor \tactics{}. \tactics{} has higher RMSE in most cells, and the
error is bias-dominated (bias$^2$/MSE near 100\%), as expected of a
deliberately unrepresentative diagnostic batch, where concentrating hard
segments makes the batch a poor estimator of the corpus average but a stronger
detector of the system ranking (Table~\ref{tab:reference_verdicts},
Figure~\ref{fig:sample_efficiency}).}
\label{tab:estimation_rmse}
\end{table}

\section{System-Ranking Recovery}
\label{app:recovery}

Table~\ref{tab:verdict_recovery} reports how reliably a sampled batch recovers
the correct system ranking, complementing the corpus-average estimation error
of Appendix~\ref{app:estimation}. Using the full-pool audit verdict of
Table~\ref{tab:reference_verdicts} as reference, for each budget we draw many
batches and run a paired $t$-test on the per-segment MQM deltas, reporting the
median $|t|$ and the \emph{recovery rate} (draws that are significant at
$p<0.05$ and match the verdict sign).

The pattern mirrors the estimation trade-off. Where the true effect is clear
(en$\rightarrow$hi, ja$\rightarrow$en) both samplers recover it almost always.
Where it is small (ja$\rightarrow$zh, $\Delta=-0.086$) \tactics{} recovers the
verdict at 1600--2400 segments (100\% vs.\ 62--93\%), and on the null control
(en$\rightarrow$fr, $\Delta=+0.016$) it raises no false verdict (0\% vs.\
random's 4--6\%).

\begin{table}[t]
\centering
\small
\setlength{\tabcolsep}{4pt}
\renewcommand{\arraystretch}{0.95}
\resizebox{\columnwidth}{!}{%
\begin{tabular}{lrrrrrr}
\toprule
Locale & Batch & $|t|_{\text{rand}}$ & $|t|_{\text{TAC}}$ & ratio
 & Rec$_{\text{rand}}$ & Rec$_{\text{TAC}}$ \\
\midrule
\multirow{5}{*}{en$\rightarrow$fr}
 & 200  & 0.75  & 1.60  & 2.14 & 4\%   & 0\%   \\
 & 400  & 0.74  & 1.11  & 1.50 & 5\%   & 0\%   \\
 & 800  & 0.70  & 0.08  & 0.12 & 6\%   & 0\%   \\
 & 1600 & 0.67  & 0.22  & 0.33 & 4\%   & 0\%   \\
 & 2400 & 0.76  & 0.45  & 0.59 & 4\%   & 0\%   \\
\midrule
\multirow{5}{*}{en$\rightarrow$hi}
 & 200  & 4.32  & 5.15  & 1.19 & 95\%  & 100\% \\
 & 400  & 5.89  & 9.60  & 1.63 & 100\% & 100\% \\
 & 800  & 8.25  & 10.70 & 1.30 & 100\% & 100\% \\
 & 1600 & 11.55 & 15.11 & 1.31 & 100\% & 100\% \\
 & 2400 & 14.13 & 17.31 & 1.22 & 100\% & 100\% \\
\midrule
\multirow{5}{*}{ja$\rightarrow$en}
 & 274  & 6.11  & 4.45  & 0.73 & 100\% & 100\% \\
 & 440  & 7.64  & 6.64  & 0.87 & 100\% & 100\% \\
 & 800  & 10.08 & 7.08  & 0.70 & 100\% & 100\% \\
 & 1600 & 14.15 & 15.18 & 1.07 & 100\% & 100\% \\
 & 2400 & 17.23 & 17.76 & 1.03 & 100\% & 100\% \\
\midrule
\multirow{5}{*}{ja$\rightarrow$zh}
 & 388  & 1.10  & 1.15  & 1.04 & 11\%  & 0\%   \\
 & 529  & 1.27  & 1.21  & 0.96 & 16\%  & 0\%   \\
 & 832  & 1.53  & 1.14  & 0.75 & 26\%  & 0\%   \\
 & 1600 & 2.15  & 3.45  & 1.60 & 62\%  & 100\% \\
 & 2400 & 2.51  & 3.38  & 1.35 & 93\%  & 100\% \\
\bottomrule
\end{tabular}}
\caption{System-ranking recovery on the audited pool. For each direction and
budget, each sampler draws batches and we run a paired $t$-test on the
per-segment MQM deltas of the two systems in
Table~\ref{tab:reference_verdicts}. The $|t|$ columns give the median over
draws, ratio $=|t|_{\text{TAC}}/|t|_{\text{rand}}$ ($>1$ favors \tactics{}),
and Rec is the fraction of draws that are significant ($p<0.05$) \emph{and}
match the full-pool verdict sign. en$\rightarrow$fr is a null control, where
lower Rec is better.}
\label{tab:verdict_recovery}
\end{table}

\section{Fidelity Diagnostics}
\label{app:fidelity_diagnostics}

\begin{figure*}[t]
    \centering
    \includegraphics[width=\textwidth, height=0.82\textheight, keepaspectratio]{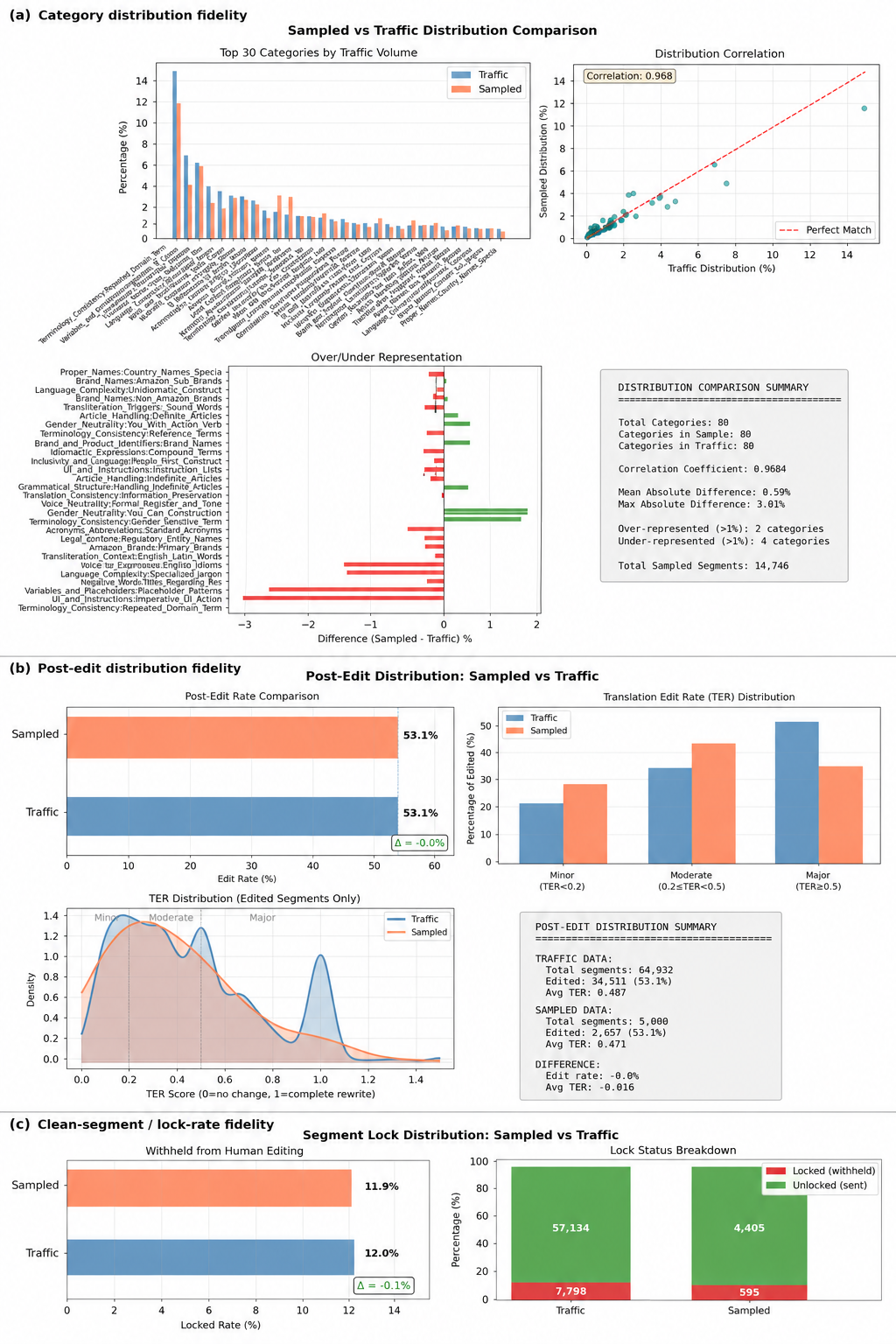}
    \caption{
    Fidelity diagnostics for the hi-IN test split.
    (a) Category distribution fidelity compares taxonomy-category
    frequencies between sampled data and the full evaluation pool.
    (b) Post-edit distribution fidelity compares edit rate and TER
    severity distributions.
    (c) Clean-segment / lock-rate fidelity compares the percentage of
    segments withheld from human editing. These diagnostics illustrate
    how Level~3 aligns the final sampled set with the full pool.
    }
    \label{fig:fidelity_diagnostics}
\end{figure*}

Figure~\ref{fig:fidelity_diagnostics} shows representative Level-3
fidelity diagnostics for the hi-IN test split. The sampled set closely
matches the full evaluation pool across taxonomy-category distribution,
post-edit distribution, and clean-segment/lock-rate statistics.

\section{Taxonomy Generation Pipeline}
\label{app:taxonomy_prompts}

The taxonomy generation pipeline consists of three LLM-based agents orchestrated in an iterative loop,
\textbf{Extract} $\rightarrow$ \textbf{Audit} $\rightarrow$ \textbf{Refine} $\rightarrow$ \textbf{Audit} $\rightarrow$ \ldots,
running for up to 5 iterations or until the auditor marks the taxonomy as complete.
Each agent can use a different LLM instance.
Below we reproduce the full system prompts (DSPy signatures) for each agent.

\subsection{Taxonomy Extraction Agent}
\label{app:extraction_agent}

The extraction agent operates in two stages, \emph{Feature Extraction} followed by \emph{Synthesis}.

\subsubsection{Stage 1: Feature Extraction}

\begin{quote}
\small
\ttfamily
Analyze a translation style guide and identify SOURCE-TEXT TRIGGERS.

\medskip
IGNORE THE FOLLOWING:\\
-- Rules that only apply to the target (e.g., `no double spaces in target')\\
-- Team-specific or project-specific instructions (e.g., ``Note for X projects'', ``Exception for Y projects'', ``Specific instructions for Z'')\\
-- Rules from other organizations/clients (only extract rules for the specified organization\_name)

\medskip
FOCUS ON:\\
-- Universal source attributes that force a specific translation decision\\
-- Linguistic markers (e.g., Second-person pronouns)\\
-- Content types (e.g., UI labels, Legal names)\\
-- Technical constraints (e.g., ICU placeholders)\\
-- General rules applicable across all projects for this organization

\medskip
\textbf{Inputs:}\\
\textnormal{\textit{markdown\_content}}: The style guide content\\
\textnormal{\textit{organization\_name}}: Organization/Client name of the style guide

\medskip
\textbf{Output:}\\
\textnormal{\textit{source\_features}}: Structured list of features including categories, names, rules, and pattern hints.
\end{quote}

\subsubsection{Stage 2: Taxonomy Synthesis}

\begin{quote}
\small
\ttfamily
Consolidates raw features into a source-centric hierarchical TaxonomyStructure.

\medskip
Logic:\\
1. PERSPECTIVE SHIFT: Transform target-evaluation rules into source-text triggers.
   (e.g., Change `Check for natural translation' to `Idiomatic/Complex Source').\\
2. BUCKETING: Group into `Deterministic' (Regex-ready) vs. `Semantic' (LLM-required).\\
3. LABELING: Ensure Semantic labels describe what to LOOK FOR in the Source-Text.\\
4. DEDUPLICATION: Merge identical source-triggers from different guide sections.

\medskip
IGNORE DURING SYNTHESIS:\\
-- Team-specific or project-specific instructions\\
-- Rules that only apply to specific teams, projects, or contexts

\medskip
\textbf{Input:}\\
\textnormal{\textit{raw\_features}}: Aggregated features from guide extraction

\medskip
\textbf{Output:}\\
\textnormal{\textit{organized\_taxonomy}}: The final nested hierarchy (Deterministic/Semantic) ready for the Classifier.
\end{quote}

\subsection{Taxonomy Auditor Agent}
\label{app:auditor_agent}

\begin{quote}
\small
\ttfamily
Audit taxonomy extraction for structural completeness.

\medskip
SCOPE CONSTRAINTS:\\
-- Be pragmatic, not exhaustive.\\
-- Focus on STRUCTURAL completeness, not exhaustive examples or word lists.\\
-- A category is complete if it captures the CONCEPT and RULE, not every instance.\\
-- Do NOT request specific word pairs, complete tables, or dictionary references.\\
-- Maximum 3--5 high-priority structural gaps per audit.\\
-- Mark is\_complete=True if >80\% of major rule categories are represented.

\medskip
IGNORE DURING AUDIT:\\
-- Team-specific or project-specific instructions\\
-- Rules from other organizations/clients\\
-- DO NOT flag team/project-specific content as ``missing'' from the taxonomy

\medskip
MARK COMPLETE (is\_complete=True) IF:\\
-- All major GENERAL category TYPES are represented\\
-- Patterns capture general structure adequately\\
-- Descriptions explain the rule/concept sufficiently\\
-- Source document lacks extractable rules (intro/purpose text only)\\
-- You cannot meaningfully validate against source content

\medskip
MARK INCOMPLETE (is\_complete=False) ONLY IF:\\
-- Entire GENERAL rule categories are clearly missing\\
-- Critical patterns have no representation at all\\
-- AND source document contains actual rules to validate against

\medskip
EDGE CASE -- AUTOMATIC PASS:\\
If source document contains only introductory/purpose statements with no actual rules, patterns, or guidelines to extract, you MUST return is\_complete=True. Do not flag items as ``missing'' when the source itself contains nothing to extract.

\medskip
LOGICAL CONSTRAINT:\\
``Cannot assess completeness'' + ``is\_complete=False'' is INVALID.
If validation is impossible due to source content $\rightarrow$ is\_complete=True automatically.
Do NOT waste iterations on unvalidatable content.

\medskip
DO NOT REQUEST:\\
-- Exhaustive word lists or substitution tables\\
-- Specific external reference links or dictionary citations\\
-- Every example from source document\\
-- Granular details requiring multiple extraction passes\\
-- Items not present in the source document\\
-- Team/project-specific rules or exceptions

\medskip
\textbf{Inputs:}\\
\textnormal{\textit{markdown\_content}}: The original style guide text\\
\textnormal{\textit{organization\_name}}: Organization/Client name\\
\textnormal{\textit{current\_taxonomy}}: Current extracted TaxonomyStructure

\medskip
\textbf{Outputs:}\\
\textnormal{\textit{missing\_rules}}: List of specific rules/structural gaps identified\\
\textnormal{\textit{is\_complete}}: Boolean --- True if $>$80\% covered, False only if major structural categories are missing
\end{quote}

\subsection{Taxonomy Refiner Agent}
\label{app:refiner_agent}

\begin{quote}
\small
\ttfamily
Update the existing category taxonomy structure based on the feedback from the auditor agent.

\medskip
INSTRUCTIONS:\\
1. Review the `current\_taxonomy'.\\
2. Incorporate the missing rules identified in `audit\_report'.\\
3. Return the FULL, updated JSON structure.\\
4. Maintain the existing Group $\rightarrow$ Subgroup Hierarchy.

\medskip
IGNORE DURING REFINEMENT:\\
-- Team-specific or project-specific instructions\\
-- Rules that only apply to specific teams, projects, or contexts\\
-- DO NOT add project-specific rules when incorporating missing rules from audit\_report

\medskip
\textbf{Inputs:}\\
\textnormal{\textit{current\_taxonomy}}: The existing TaxonomyStructure JSON\\
\textnormal{\textit{audit\_report}}: Specific missing rules or errors found by the auditor

\medskip
\textbf{Output:}\\
\textnormal{\textit{revised\_taxonomy}}: The updated version of the taxonomy
\end{quote}

\subsection{Orchestration}
\label{app:orchestration}

The three agents are composed in a LangGraph workflow with conditional routing:

\begin{enumerate}
    \item \textbf{Extraction node} (iteration 0): Runs the two-stage Taxonomy Extraction Agent (Feature Extraction $\rightarrow$ Synthesis) to produce an initial \texttt{TaxonomyStructure}.
    \item \textbf{Audit node}: Runs the Taxonomy Auditor Agent against each style guide document. If $>$80\% of major rule categories are covered, marks \texttt{is\_complete=True}.
    \item \textbf{Routing decision}: If complete or \texttt{iterations} $>$ \texttt{max\_iterations} (default 5), the workflow terminates. Otherwise, it proceeds to refinement.
    \item \textbf{Refinement node} (iterations $\geq$ 1): Runs the Taxonomy Refiner Agent to incorporate the auditor's missing rules into the existing taxonomy, then loops back to the audit node.
\end{enumerate}

Each stage supports a separate LLM instance (\texttt{extraction\_lm}, \texttt{audit\_lm}, \texttt{refinement\_lm}), enabling heterogeneous model configurations (e.g., a stronger model for extraction, a faster model for auditing).
All modules use DSPy's \texttt{ChainOfThought} wrapper, which elicits step-by-step reasoning before producing the structured output.

\section{Taxonomy Validation}
\label{app:taxonomy_validation}

We assess taxonomy consistency via cross-family agreement between two
independent classifiers from different model families, Kimi K2.5 and Claude
Haiku 4.5. Agreement improves after refinement (Cohen's $\kappa$ from 0.27 to
0.51 on a 100-segment pilot), indicating clearer category boundaries. A
taxonomy spot checker additionally samples induced categories and verifies
that each is grounded in a prescriptive style-guide rule, clearly defined,
and distinguishable from its neighbors, and Table~\ref{tab:sg_taxonomy_mapping}
shows this mapping for fr-FR.

Taxonomy sizes range from 48 to 177 categories across locales, reflecting
variation in style-guide complexity. English variants (e.g., en-AU, en-IE,
en-ZA) are largest (128--177), East Asian locales (ja-JP, ko-KR, zh-TW) most
compact (76--87), and Romance locales fall in between (77--150).

\begin{table*}[t]
\centering
\footnotesize
\setlength{\tabcolsep}{3pt}
\renewcommand{\arraystretch}{0.92}
\begin{tabular}{p{2.7cm} p{3.1cm} p{1.1cm} p{5.4cm}}
\toprule
\textbf{Style Guide Rule} & \textbf{Taxonomy Category} & \textbf{Type} & \textbf{Spot-Check Example} \\
\midrule
\multicolumn{4}{c}{\textit{Fluency -- Grammar \& Spelling}} \\
\midrule

No plural on acronyms (PDFs $\rightarrow$ PDF)
 & Acronyms\_Abbrev. / Pluralized\_Acronyms
 & Det.
 & \texttt{\textbackslash b[A-Z]\{2,\}s\textbackslash b} catches ``PDFs'', ``DVDs'' \\

Country abbrev.\ (UK $\rightarrow$ Royaume-Uni)
 & Acronyms\_Abbrev. / Country\_Abbreviations
 & Det.
 & \texttt{UK|U.K.} $\rightarrow$ ``Royaume-Uni'' \\

Dotted abbrev.\ (U.S. $\rightarrow$ \'Etats-Unis)
 & Acronyms\_Abbrev. / Dotted\_Abbreviations
 & Det.
 & \texttt{U.S.} matched by \texttt{[A-Z]\textbackslash .([A-Z]\textbackslash .)+} \\

Tech spelling (Wi-Fi, e-mail)
 & Technology\_Terms / Tech\_Spelling
 & Det.
 & ``WiFi'' $\rightarrow$ ``Wi-Fi''; ``email'' $\rightarrow$ ``e-mail'' \\

Web capitalization
 & Technology\_Terms / Web\_Capitalization
 & Det.
 & ``web'' $\rightarrow$ ``Web'' \\

Proper names invariable (iPads)
 & Proper\_Names / Apple\_Products
 & Det.
 & ``iPads'' $\rightarrow$ ``les iPad'' (no plural) \\

\midrule
\multicolumn{4}{c}{\textit{Fluency -- Typography}} \\
\midrule

Ampersand $\rightarrow$ ``et''
 & Punctuation\_Typo. / Ampersand
 & Det.
 & ``Shoes \& Bags'' $\rightarrow$ ``Chaussures et sacs'' \\

Hashtag $\rightarrow$ n\textsuperscript{o}
 & Punctuation\_Typo. / Hashtag\_Number
 & Det.
 & ``\#1'' $\rightarrow$ ``n\textsuperscript{o} 1'' \\

No exclamation in UI
 & Punctuation\_Typo. / Exclamation\_in\_UI
 & Det.
 & ``Welcome!'' $\rightarrow$ remove ``!'' in UI strings \\

\midrule
\multicolumn{4}{c}{\textit{Locale Convention}} \\
\midrule

Currency format (symbol after number)
 & Currency\_Pricing / Currency\_Symbols
 & Det.
 & ``\$19.99'' $\rightarrow$ ``19,99 \$'' or ``19,99 USD'' \\

Date format (MM/DD $\rightarrow$ DD/MM)
 & Date\_Formats / US\_Numeric\_Date
 & Det.
 & ``7/14/2013'' $\rightarrow$ ``14/07/2013'' \\

Written date reorder
 & Date\_Formats / Written\_Month\_Date
 & Det.
 & ``July 14, 2013'' $\rightarrow$ ``14 juillet 2013'' \\

12h $\rightarrow$ 24h time
 & Time\_Formats / Twelve\_Hour\_Time
 & Det.
 & ``8:30 pm'' $\rightarrow$ ``20 h 30'' \\

Thousands sep.\ (comma $\rightarrow$ space)
 & Number\_Formats / Thousands\_Separator
 & Det.
 & ``2,935,000'' $\rightarrow$ ``2 935 000'' \\

Decimal sep.\ (period $\rightarrow$ comma)
 & Number\_Formats / Decimal\_Numbers
 & Det.
 & ``11.5'' $\rightarrow$ ``11,5'' \\

Percent spacing
 & Number\_Formats / Percentage\_Values
 & Det.
 & ``10\%'' $\rightarrow$ ``10 \%'' \\

Discount format
 & Number\_Formats / Discount\_Expressions
 & Det.
 & ``40\% off'' $\rightarrow$ ``- 40 \%'' \\

Fahrenheit $\rightarrow$ Celsius
 & Measurements / Temperature\_Fahrenheit
 & Det.
 & ``72\textdegree F'' $\rightarrow$ ``22 \textdegree C'' \\

Imperial $\rightarrow$ metric
 & Measurements / Imperial\_Units
 & Det.
 & ``12 inches'' $\rightarrow$ ``30,48 cm'' \\

Clothing sizes
 & Measurements / Clothing\_Sizes
 & Det.
 & ``Small'' $\rightarrow$ ``Taille S'' \\

\midrule
\multicolumn{4}{c}{\textit{Style}} \\
\midrule

Polite ``vous'' form
 & Voice\_Tone / Second\_Person\_Address
 & Sem.
 & ``you'' $\rightarrow$ always ``vous'' (formal) \\

Imperative in instructions
 & Instruction\_Style / Imperative
 & Sem.
 & ``Print label'' $\rightarrow$ ``Imprimez l'\'etiquette'' \\

Infinitive for buttons
 & Instruction\_Style / Button\_Link\_Labels
 & Sem.
 & ``Find a package'' $\rightarrow$ ``Trouver un paquet'' \\

Gender-neutral roles
 & Gender\_Neutrality / Role\_Terms
 & Sem.
 & ``specialist'' $\rightarrow$ ``Sp\'ecialiste'' (not ``Expert(e)'') \\

Predicate adj.\ with ``you''
 & Gender\_Neutrality / Predicate\_Adj.
 & Sem.
 & ``you're ready'' $\rightarrow$ rephrase to avoid gendered agreement \\

Long sentences
 & Language\_Complexity / Long\_Complex
 & Sem.
 & Break up multi-clause sentences for readability \\

\bottomrule
\end{tabular}
\caption{Spot-check mapping between French (fr-FR) localization style guide rules and taxonomy categories.
\textbf{Type}: Det.\ = Deterministic (regex-based), Sem.\ = Semantic (LLM-based).
Each row shows a style guide rule, its corresponding taxonomy category, detection type, and a concrete example of how the rule is operationalized.}
\label{tab:sg_taxonomy_mapping}
\end{table*}

\section{Example Taxonomy for French (fr-FR)}
\label{app:taxonomy_example}

\begin{figure*}[t]
    \centering
    \includegraphics[width=0.95\textwidth]{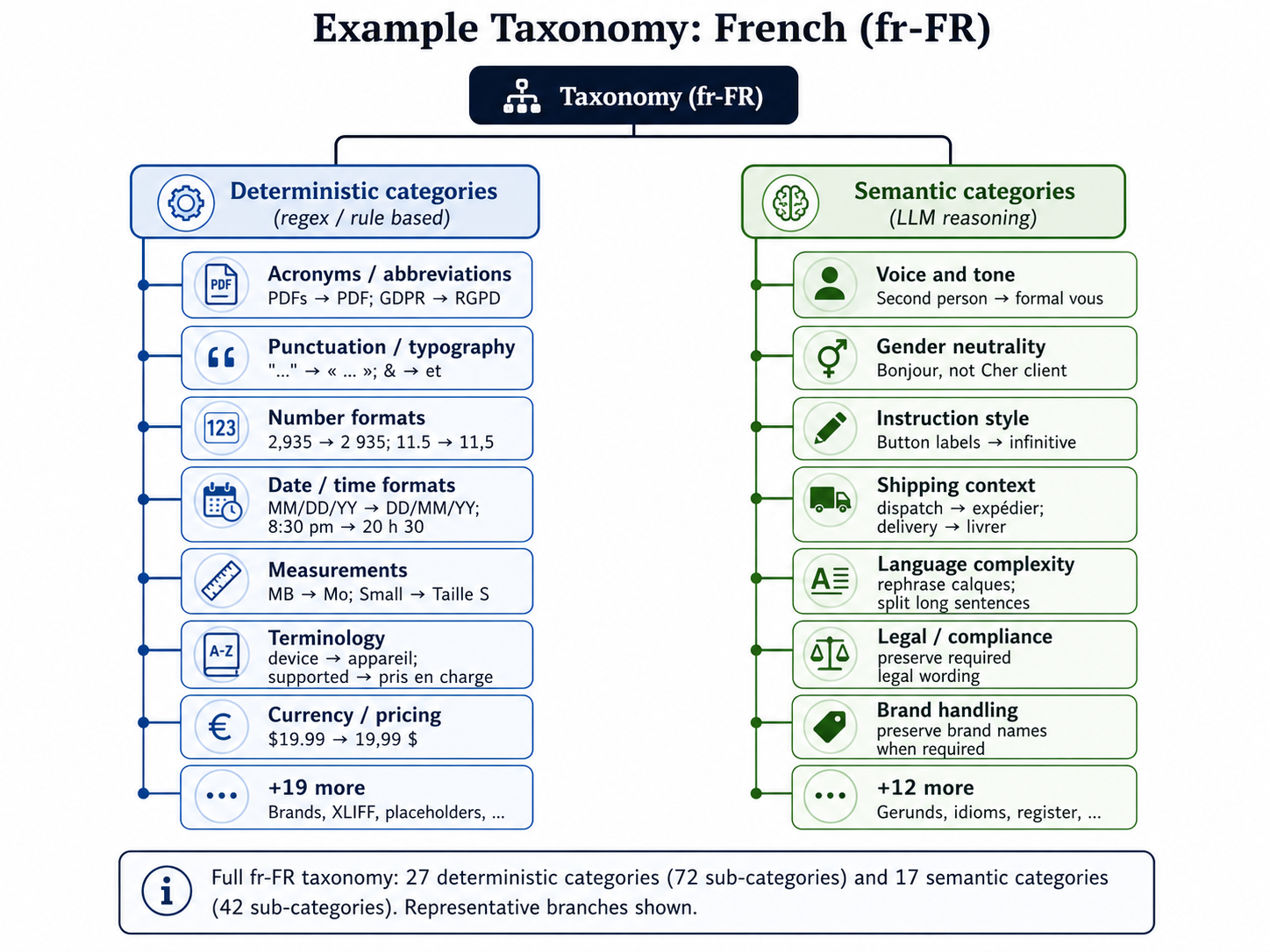}
    \caption{Condensed example taxonomy for French (fr-FR). Deterministic
    categories capture rule-matchable phenomena such as punctuation,
    number formats, terminology, and currency conventions. Semantic
    categories capture context-dependent phenomena such as tone, gender
    neutrality, instruction style, and shipping context.}
    \label{fig:taxonomy_tree}
\end{figure*}

Figure~\ref{fig:taxonomy_tree} shows a condensed view of the taxonomy induced
for French (fr-FR), separating \textit{Deterministic} categories (detectable
with regex-style or rule-based patterns) from \textit{Semantic} categories
(requiring context-aware LLM reasoning). The full taxonomy contains 27
deterministic categories (72 sub-categories) and 17 semantic categories
(42 sub-categories).

\label{sec:appendix}

\end{document}